\documentclass[letterpaper]{article} % DO NOT CHANGE THIS
\usepackage{aaai25}  % DO NOT CHANGE THIS
\usepackage{times}  % DO NOT CHANGE THIS
\usepackage{helvet}  % DO NOT CHANGE THIS
\usepackage{courier}  % DO NOT CHANGE THIS
\usepackage[hyphens]{url}  % DO NOT CHANGE THIS
\usepackage{graphicx} % DO NOT CHANGE THIS
\usepackage{natbib}  % DO NOT CHANGE THIS AND DO NOT ADD ANY OPTIONS TO IT
\usepackage{caption} % DO NOT CHANGE THIS AND DO NOT ADD ANY OPTIONS TO IT
\usepackage{algorithm}
\usepackage{algorithmic}

\usepackage{amssymb}
\usepackage{amsmath}
\usepackage{multirow}
\usepackage[table,xcdraw]{xcolor}

\usepackage{newfloat}
\usepackage{listings}
\DeclareCaptionStyle{ruled}{labelfont=normalfont,labelsep=colon,strut=off} % DO NOT CHANGE THIS
\floatstyle{ruled}
\newfloat{listing}{tb}{lst}{}
\floatname{listing}{Listing}
\title{DuSSS: Dual Semantic Similarity-Supervised Vision-Language Model for Semi-Supervised Medical Image Segmentation}
\author{
    Qingtao Pan\textsuperscript{\rm 1,\rm 2},
    Wenhao Qiao\textsuperscript{\rm 1,\rm 2},
    Jingjiao Lou\textsuperscript{\rm 1,\rm 2},
    Bing Ji\textsuperscript{\rm 1,\rm 2}\thanks{Corresponding Author},
    Shuo Li\textsuperscript{\rm 3}
}
\affiliations{
    \textsuperscript{\rm 1}School of Control Science and Engineering, Shandong University, Jinan, China\\
    \textsuperscript{\rm 2}Key Laboratory of Machine Intelligence and System Control, Ministry of Education, China\\
    \textsuperscript{\rm 3}Department of Computer and Data Science and Department of Biomedical Engineering, \\ Case Western Reserve University, USA\\
    \{qingtaopan33, wenhaoo.qiao, jingjiaolou, slishuo\}@gmail.com, b.ji@sdu.edu.cn
}

\usepackage{bibentry}
\begin{document}

\maketitle

\begin{abstract}
Semi-supervised medical image segmentation (SSMIS) uses consistency learning to regularize model training, which alleviates the burden of pixel-wise manual annotations. However, it often suffers from error supervision from low-quality pseudo labels. Vision-Language Model (VLM) has great potential to enhance pseudo labels by introducing text prompt guided multimodal supervision information. It nevertheless faces the cross-modal problem: the obtained messages tend to correspond to multiple targets. To address aforementioned problems, we propose a Dual Semantic Similarity-Supervised VLM (DuSSS) for SSMIS. Specifically, \textbf{1)} a Dual Contrastive Learning (DCL) is designed to improve cross-modal semantic consistency by capturing intrinsic representations within each modality and semantic correlations across modalities. \textbf{2)} To encourage the learning of multiple semantic correspondences, a Semantic Similarity-Supervision strategy (SSS) is proposed and injected into each contrastive learning process in DCL, supervising semantic similarity via the distribution-based uncertainty levels. Furthermore, a novel VLM-based SSMIS network is designed to compensate for the quality deficiencies of pseudo-labels. It utilizes the pretrained VLM to generate text prompt guided supervision information, refining the pseudo label for better consistency regularization. Experimental results demonstrate that our DuSSS achieves outstanding performance with Dice of 82.52\%, 74.61\% and 78.03\% on three public datasets (QaTa-COV19, BM-Seg and MoNuSeg).
\end{abstract}

% Uncomment the following to link to your code, datasets, an extended version or similar.
%
\begin{links}
    \link{Code}{https://github.com/QingtaoPan/DuSSS/}
\end{links}

\section{Introduction}
\label{sec1}

Medical image segmentation aims to divide medical images into specific regions with unique attributes. It contributes to detecting abnormal areas and providing clinical guidance \cite{refa1}. In clinical practice, achieving precise segmentation results necessitates manual implementation and there is an urgent need for automatic medical image segmentation to aid in diagnosis and treatment. Fully-supervised medical image segmentation methods, such as U-Net \cite{refa2} and its variants \cite{refa3,refa4,refa5}, have been developed for medical image segmentation through an encoder and a decoder in a U-shaped architecture, connected by skip connections. However, fully-supervised methods need a large amount of pixel-level annotated data for model training and labeling such pixel-level annotations is laborious and requires expert knowledge especially in medical images, resulting in that labeled data are expensive or simply unavailable \cite{ref67}. Semi-supervised medical image segmentation (SSMIS) is a method that utilizes a small amount of labeled data and a large amount of unlabeled data to learn segmentation models \cite{refa7}.

\begin{figure}[t]
\centering
\includegraphics[width=\linewidth]{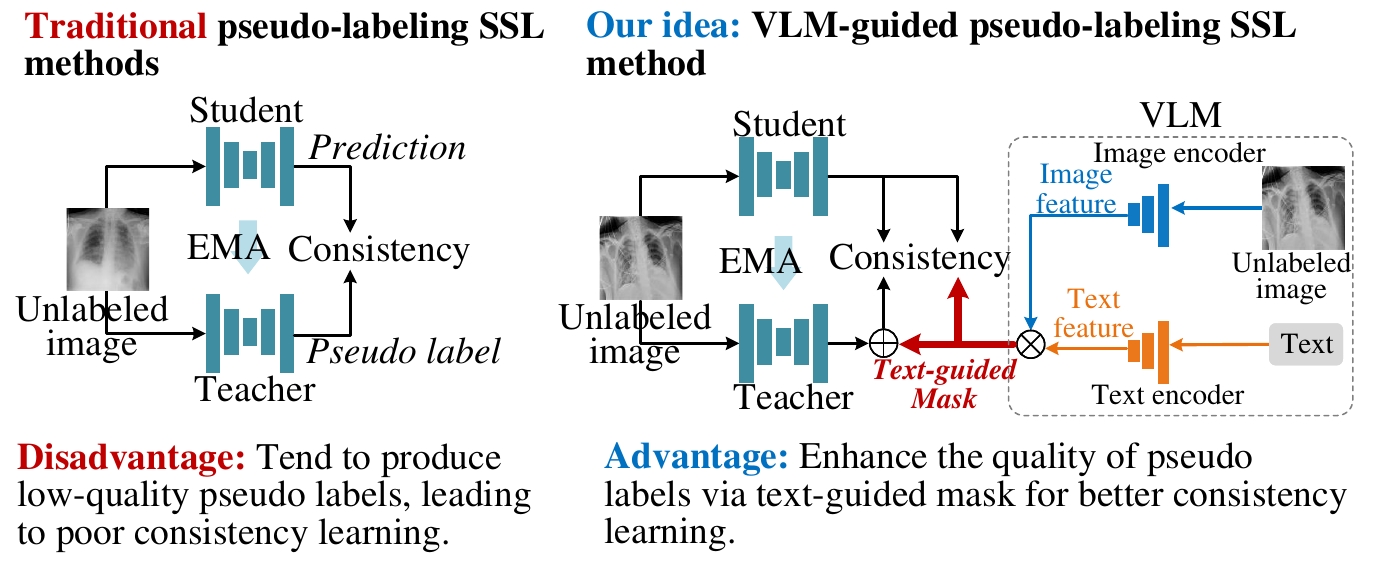}
\caption{The VLM has potential to enhance pseudo labels via text-guided mask, improving consistency learning.}
\label{fig1}
\end{figure}

SSMIS methods, such as \cite{refa8,ref9,refa11,ref64}, have shown promising performance for semi-supervised segmentation by combining consistency regularization and pseudo labeling via cross supervision between the sub-networks. However, one key disadvantage of these approaches is that they may generate low-quality pseudo labels \citep{ref59}, which misleads model training and causes the model to learn incorrect features, thus failing to effectively utilize unlabeled data and disrupting consistency learning on unlabeled data. Therefore, the question that comes to mind is: how to effectively improve the quality of pseudo labels for SSMIS.

\begin{figure}[t]
\centering
\includegraphics[width=\linewidth]{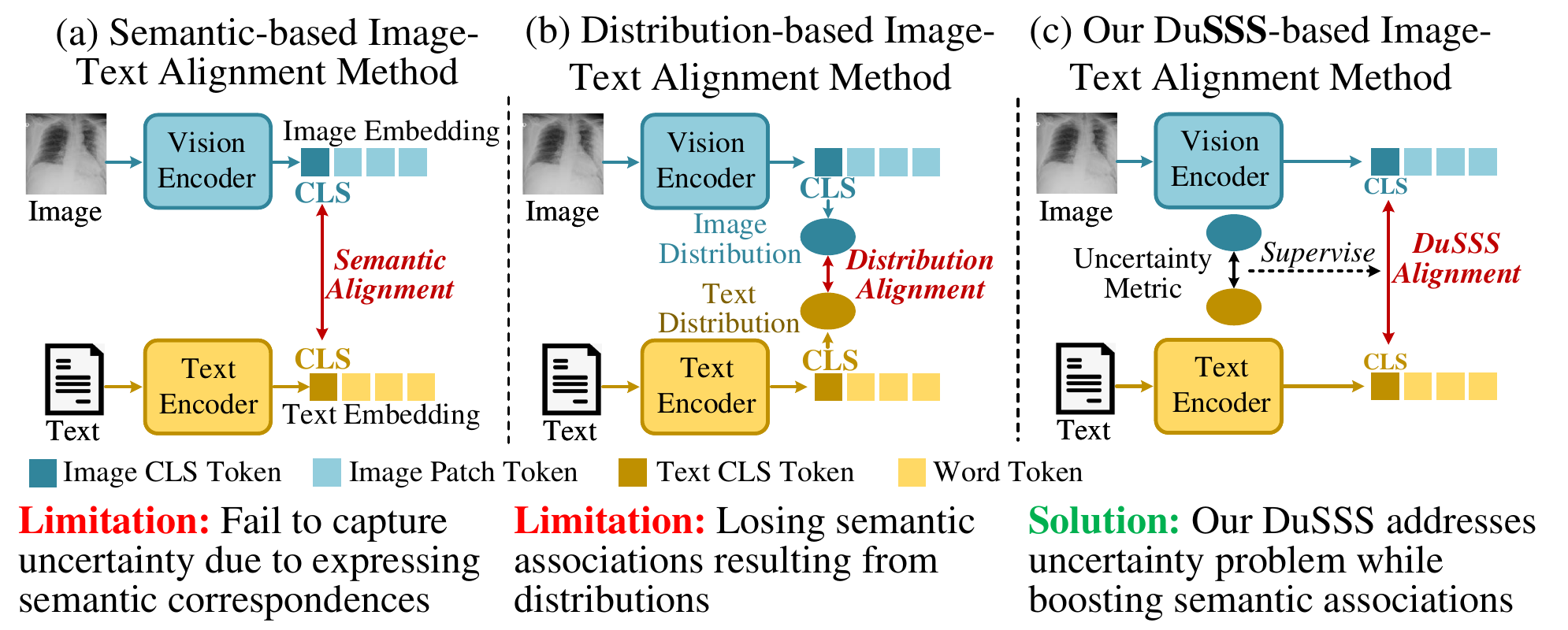}
\caption{The limitations of current VLM methods and the solution of our DuSSS.}
\label{fig2}
\end{figure}

Vision-Language Model (VLM) has great potential to enhance the quality of pseudo-labels. It produces text-guided supervision information by leveraging textual prompts to describe visual content, thus enhancing pseudo labels for various tasks \citep{refa12}, as shown in Fig. 1. Once successful, this approach will significantly improve the performance of SSMIS by guiding the segmentation model to locate target segmentation regions with textual prompts. Although promising performance of current VLMs, such as CLIP \citep{ref25}, MedCLIP \citep{refa14}, MGCA \citep{refa15}, etc, the cross-modal uncertainty remains a significant problem since they merely conduct one-to-one alignment between image and text (Fig. \ref{fig2} (a)). Specifically, multiple images/texts may correspond to one text/image, which manifests cross-modal uncertainty.

Utilizing distribution to represent semantic embeddings is a prominent approach for uncertainty awareness \cite{ref1,ref2,ref4}. This approach commonly transfers semantic embeddings as distribution representations for image-text alignment through Gaussian modeling \cite{ref6,ref7}, to learn the uncertainty (Fig. \ref{fig2} (b)). The distribution's variance is computed for uncertainty judgment. Although the variance reflects distribution discrepancies, relying solely on distribution representations lose original semantic attributes, leading to poor semantic associations between image and text. Therefore, we argue that it is necessary to address the uncertainty problem while detaining semantic attributes.

In this paper, we propose a novel VLM paradigm with Dual Semantic Similarity-Supervision (DuSSS) to learn multiple semantic correspondences (Fig. \ref{fig2} (c)). Specifically, \textbf{1)} a Dual Contrastive Learning (DCL) is constructed for cross- and intra-modal contrastive learning. It reinforces semantic correlations across modalities and captures implicit representation relationships within each modality, thereby prompting image-text alignment. \textbf{2)} A Semantic Similarity-Supervision strategy (SSS) is proposed to supervise semantic similarity based on corresponding uncertainty levels by incorporated it into each contrastive learning process in DCL, learning potential uncertainty correspondences across modalities and within modalities. \textbf{3)} For the first time, a text-guided segmentation network is constructed to compensate for quality deficiencies of pseudo-labels for SSMIS.

Our contributions are summarized as follows:

\begin{itemize}

    \item This is the first VLM-based method for pseudo labeling SSMIS. It compensates for the quality deficiencies of pseudo-labels by utilizing the advantages of text prompts to locate segmentation regions.
    
    \item A new SSS supervises semantic similarity via uncertainty levels. It promotes model's understanding of data pairs that are similar yet semantically ambiguous, mitigating impacts of uncertain correspondences.
    
    \item A novel DCL boosts semantic associations across modalities and captures intrinsic representation relationships within each modality, thereby boosting cross-modal semantic consistency.
    
    \item Extensive experiments are conducted on three public medical image segmentation datasets. Comprehensive results demonstrate the effectiveness of each component of our method and the superiority of our DuSSS over the state-of-the-art methods.
\end{itemize}

\section{Related Works}
\label{sec2}

\subsection{Semi-Supervise Medical Image Segmentation}
In the field of SSMIS, the main methods includes self-training methods \citep{refa16,refa17}, adversarial training methods \citep{refa18,refa19}, co-training methods \citep{refa11,refa21}, and consistency regularization methods \citep{refa22,refa23}. Consistency regularization focuses on maintaining consistent model predictions under different perturbations. The state-of-the-art technique is Mean Teacher (MT) \citep{ref62}. In MT, the teacher model is employed to generate pseudo-labels for unlabeled data while maintaining prediction consistency between the teacher and student models through various regularization methods. Afterward, the teacher model is the exponential moving average (EMA) of the student model's weights. This method enables the teacher model to continually aggregate historical prediction information from unlabeled data. Subsequent improvements use different consistency regularization strategies to improve the prediction quality of unlabeled data \citep{refa8,refa26}.  However, these methods are still based on the single-modal approaches, leading to poor pseudo labels. In this paper, we introducing VLM into SSMIS to enhancing the quality of pseudo labels by generating text prompt guided multimodal supervision information.

\subsection{Vision-Language Model}
Although existing VLMs learns generic visual-textual representations by aligning image-text, they are limited by uncertainty awareness of cross-modal and intra-modal. CLIP \cite{ref25} is a representative work of VLM, using the contrastive loss to calculate similarity scores between images and texts. ViLT \cite{ref26} is a more efficient architecture that deals with visual feature using interaction layers. Accordingly, numerous works use text information to improve the image segmentation capabilities \cite{ref28,ref30,ref32}. For instance, Yang et al. \cite{ref28} conducted early fusion of image-text features in intermediate layers of a transformer network, achieving significantly cross-modal alignment. Ding et al. \cite{ref30} built an encoder decoder attention network with transformer and multi-head attention to provide the language expression ”queries” for the given image. Inspired by VLM in natural images, a few works have started utilizing text information for medical image analysis \cite{ref34,ref35}. Li et al. \cite{ref36} proposed a Language meets Vision Transformer model (LViT) to incorporated text annotations with images in down-sampling and up-sampling processes, compensating for the quality deficiencies in image data. Boecking et al. \cite{ref36-2} used the attention weights learned during local alignment to conduct medical semantic segmentation. In this work, we introduce cross-modal and intra-modal self-supervision with uncertainty awareness for better image-text alignment.

\section{Methodology}
\label{sec3}

\begin{figure*}[t]
\centering
\includegraphics[width=\linewidth]{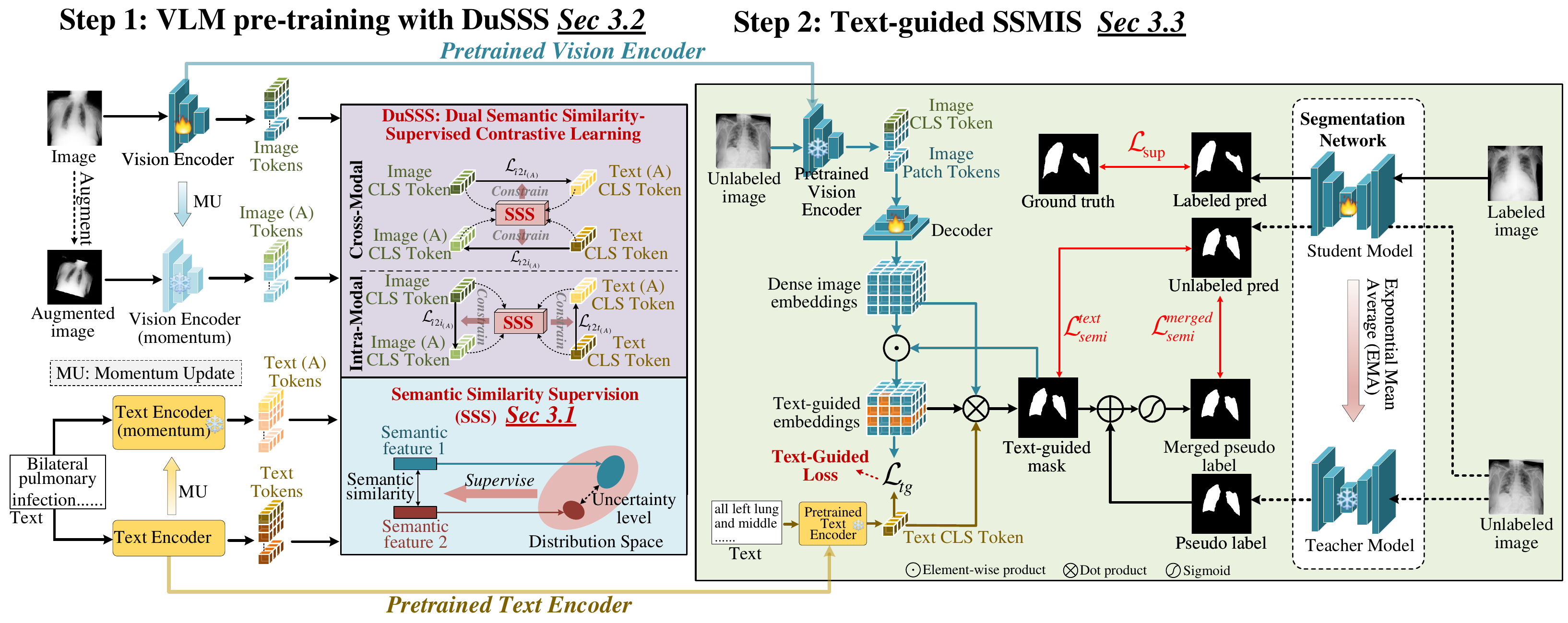}
\caption{The framework of our DuSSS driven VLM for SSMIS. \textbf{Step 1:} Our DuSSS improves the ability of uncertainty understanding in VLM pre-training, thus enhancing the model's robustness for image-text alignment. \textbf{Step 2:} The text-guided SSMIS improves the quality of pseudo-labels for reliable semi-supervised consistency learning.}
\label{fig3}
\end{figure*}

Our DuSSS (Fig. \ref{fig3}) proposes uncertainty-aware VLM for SSMIS. Specifically, it comprises two steps. \textbf{Step 1: VLM pre-training with DuSSS}. The DuSSS promotes cross-modal semantic associations via DCL and supervises semantic similarity based on uncertainty levels during contrastive learning process for uncertainty learning. It addresses uncertainty problems across modalities. \textbf{Step 2: Text-guided SSMIS}. It generates multimodal supervision information (i.e., text-guided mask) to improve pseudo-label quality.

\subsection{SSS for Uncertain Correspondence Learning}
The SSS supervises semantic similarity based on the corresponding uncertainty levels to perceive uncertain correspondences. It comprehends intricate semantic relationships between data pairs that are similar yet semantically ambiguous, improving semantic consistency representations.

The distribution distance is regarded as the uncertainty level, which supervises semantic similarity measured by Euclidean distance. For a paired data $(x_1,x_2)$, their semantic similarity is defined as $D_s(x_1,x_2)=\left\Vert s_1-s_2 \right\Vert_2$. ($s_1$, $s_2$) are the semantic embeddings of $(x_1,x_2)$. To measure uncertainty level, semantic embeddings are transformed to multivariate Gaussian distributions. The mean vector $\mu$ and variance vector $\sigma^2$ denote the center position and the distribution scope, respectively. The 2-Wasserstein \cite{ref50,ref26} is utilized to measure the difference (i.e. uncertainty level) of multivariate Gaussian distributions. The following defines the 2-Wasserstein of two Gaussian distributions $\mathcal{N}(\mu_{1},\sigma_{1})$ and $\mathcal{N}(\mu_{2},\sigma_{2})$.

\begin{equation}
D_{2W} = ||\mu_1-\mu_2||_2^2+||\sigma_1-\sigma_2||_2^2
\end{equation}
where $\mu$ represents the mean vector and $\sigma$ is the standard deviation vector. The global information of [CLS] is utilized to model the uncertainty distribution. The uncertainty level between the paired data $(x_1,x_2)$ is given by:

\begin{equation}
D_{u}(x_1,x_2)=a\cdot D_{2W}(x_{1{[\mathtt{CLS}]}},x_{2{[\mathtt{CLS}]}})+b
\end{equation}
where $a$ is a positive scale that controls the degree of uncertainty and $b$ is a deviation value. When comparing two data, both the semantic similarity $D_s(\cdot)$ and the uncertainty level $D_u{(\cdot)}$ need to be considered. The semantic similarity is constrained when large uncertainty exists. Here, the ratio of the uncertainty level to the semantic similarity is used to define the relative uncertainty $\widehat{D}_{u}$. The SSS is defined as:

\begin{equation}
\widehat{D}_{u}(x_1,x_2)=\frac{D_{u}(x_1,x_2)}{D_{s}(x_1,x_2)}
\end{equation}

\begin{equation}
D_{SSS}(\mathbf{x}_1,\mathbf{x}_2)=e^{-{\lambda{\widehat{D}_{u}}(\mathbf{x}_1,\mathbf{x}_2)}}
\end{equation}
where $\lambda$ is a parameter for controlling constraint degree.

\textbf{Summarized Advantages:} The SSS provides a novel approach to handle uncertainty for robust image-text alignment. It supervises semantic similarity via the uncertainty levels measured by distribution representations. Therefore, it promotes the ability of similarity judgments for ambiguous sample pairs. It addresses the problem in existing distribution-based VLM methods where original semantic attributes are losing.

\subsection{VLM Pre-training with DuSSS}
The DuSSS learns multiple semantic correspondences across modalities and within each modality for uncertainty learning. Specifically, the DCL is conducted to advance the semantic correlation between images and texts by driving cross- and intra-modal representation learning. The SSS is integrated into cross-modal and intra-modal contrastive learning in DCL to supervise semantic similarity in each contrastive learning process via uncertainty levels, solving semantic uncertainty.

\textbf{1) SSS-based Cross-Modal Contrastive Learning (CMC)} aims to learn multiple semantic correspondences between image and text embeddings. That is, pull the matched image-text embeddings together and push the unmatched image-text embeddings away. To avoid potential uncertainties, the cosine similarity $sim(I,T)$ between image and text is constrained via $D_{SSS}(I,T)$, thus obtaining the uncertainty cosine similarity $s\widehat{i}m(I,T)$. The InfoNCE loss \cite{ref54} is used to optimize $s\widehat{i}m(I,T)$, thus enabling the positive image-text pairs similar and the negative pairs dissimilar. For $N$ data pairs in a batch, $N$ positive image-text pairs are matched and there are $N(N-1)$ negative image-text pairs. The InfoNCE loss for the SSS-based image-to-text is denoted as:

\begin{equation}
s\widehat{i}m(I,T)=1-(1-sim(I,T)) \cdot D_{SSS}(I,T)
\end{equation}

\begin{equation}
\mathcal{L}^{I2T}_{nce}=-\mathbb{E}_{(I,T)}\left[log\frac{exp(s\widehat{i}m((I_1,T_+)/\tau)}{\sum_{n=1}^Nexp(s\widehat{i}m(I_1,\widehat{T}_n)/\tau)}\right]
\end{equation}
where $\tau$ is a learned temperature hyper-parameter. $T_+$ denotes the positive text matched with $I_1$ and $\widehat{T}=\{\widehat{T}_1,\dots,\widehat{T}_N\}$ are the negative texts that do not match $I_1$. Similarly, the InfoNCE loss for text-to-image is given by:

\begin{equation}
\mathcal{L}^{T2I}_{nce}=-\mathbb{E}_{(T,I)}\left[log\frac{exp(s\widehat{i}m(T_1,I_+)/\tau)}{\sum_{n=1}^Nexp(s\widehat{i}m(T_1,\widehat{I}_n)/\tau)}\right]
\end{equation}
where $I_+$ is the positive image that matches to $T_1$ and $\widehat{I}=\{\widehat{I}_1,\dots,\widehat{I}_N\}$ are the negative images that do not match $T_1$. Overall, we define the cross-modal loss of image-text as:

\begin{equation}
\mathcal{L}_{cmc}=\frac{1}{2}\left[\mathcal{L}^{I2T}_{nce}(I_1,T_+,\widehat{T}) + \mathcal{L}^{T2I}_{nce}(T_1,I_+,\widehat{I})\right]
\end{equation}
The CMC aligns images and texts through minimizing $\mathcal{L}_{cmc}$. However, cross-modal alignment fails to build semantic associations within each modality. To solve this, SSS-based intra-modal contrastive learning is introduced.

\textbf{2) SSS-based Intra-Modal Contrastive Learning (IMC)} captures the underlying associations among different samples within each modality. Meanwhile, the SSS is injected into IMC for weakening semantic uncertainty within each modality. For the raw image-text $(I_1,T_1)$ and the randomly augmented image-text $(I_2,T_2)$, similar to cross-modal loss, intra-modal contrastive loss is defined as:

\begin{equation}
\mathcal{L}_{imc}=\frac{1}{2}\left[\mathcal{L}^{I2I}_{nce}(I_1,I_2,\widehat{I}) + \mathcal{L}^{T2T}_{nce}(T_1,T_2,\widehat{T})\right]
\end{equation}
where $I_2$ is the image to be matched with $I_1$ and $\widehat{I}=\{\widehat{I}_1,\dots,\widehat{I_N}\}$ are the negative images. Similarly, $T_2$ is the text to be matched with $T_1$ and $\widehat{T}= \{\widehat{T}_1,\dots,\widehat{T_N}\}$ are the negative texts.

\textbf{Summarized Advantages:} The DuSSS provides a new scheme to tackle uncertainty problems across modalities. It incorporates SSS into each contrastive learning process in DCL to supervise semantic similarity via the corresponding uncertainty levels, addressing cross-modal uncertainty.

\subsection{Text-Guided SSMIS for Pseudo-Label Quality Enhancement}
The text-guided SSMIS enhances the quality of pseudo-labels via the benefit of text descriptions. The goal of SSMIS is to train a model by utilizing a labeled training dataset $X_l=\{(x_{li},y_{li})\}_i^{N_l}$ and an unlabeled training dataset $X_u=\{(x_{uj})\}_i^{N_u}$. Our text-guided SSMIS consists of two paths: 1) Text-guided pseudo-label generation, and 2) Teacher-Student Network-guided pseudo-label generation. These two paths are detailed as follows.

\textbf{1) Text-Guided Pseudo-Label Generation.}
The text-guided pseudo-label leverages the advantage of text to localize target segmentation regions. The patch-level image features $v_f^{patch}$ and text features $t_f$ from the pre-trained VLM are used to synthetize the text-guided mask. Specifically, the patch-level image features $v_f$ are decoded into pixel-level features through a grounding decoder $f_g(\cdot)$. The text-guided mask $y_u^{text}$ is calculated by dot product between $v_f$ and $t_f$, and the process is as follows.

\begin{equation}
v_f=f_g(v_f^{patch}), \quad y_u^{text}=\sigma(t_f^{\top} v_f)
\end{equation}
where $\sigma$ is Sigmoid function, and $\top$ is transposition. 

\textbf{Text-Guided Loss.} To suppress the irrelevant text guided mask generation, a text-guided loss $\mathcal{L}_{tg}$ is designed. It helps the model better understand both the global structure and local details. The text-guided loss is denoted as:

\begin{equation}
\begin{aligned}
\mathcal{L}_{tg}(v_f^t,t_f)=& -\frac12\mathbb{E}_{(f,t)}\left[log\frac{exp(sim(v_f^t,t_f))/\tau)}{\sum_{k=1}^Kexp(sim(v_f^t,\widehat{t}_{f_k})/\tau)}\right] \\ 
& - \frac12\mathbb{E}_{(t,f)}\left[log\frac{exp(sim(t_f,v_f^t))/\tau)}{\sum_{k=1}^Kexp(sim(t_f,\widehat{v}_{f_k}^t)/\tau)}\right]
\end{aligned}
\end{equation}

\textbf{2) Teacher Student Network-Guided Pseudo-Label Generation.}
The teacher and student segmentation networks have identical architecture. The student network $f_{\theta_s}$ is parameterized by $\theta_s$, while the teacher network $f_{\theta_t}$ is updated by the Exponential Moving Average (EMA) of the student. The updating process is as follows:

\begin{equation}
\theta_t=\alpha\theta_t+(1-\alpha)\theta_s
\end{equation}
where $\alpha\in$[0,1] controls the updating pace. The student network predicts the labeled image $x_l$ and the unlabeled image $x_u$. The teacher network generates the pseudo-label $y_u^{t}$ of the unlabeled image. The calculation process is:

\begin{equation}
y_l=f_{\theta_s}(x_l),\quad y_u^s=f_{\theta_s}(x_u),\quad y_u^t=f_{\theta_t}(x_u)
\end{equation}
where $y_l$ and $y_u^s$ are labeled and unlabeled prediction.

The supervised loss $\mathcal{L}_{sup}$ and semi-supervised loss $\mathcal{L}_{semi}$ are utilized to jointly optimize the segmentation network. 

For labeled images, $\mathcal{L}_{sup}$ is computed between the ground truth $y_{gt}$ and the labeled prediction $y_l$. For unlabeled images $x_u$, $\mathcal{L}_{semi}$ is obtained by $\mathcal{L}_{semi}^{merged}$ and $\mathcal{L}_{semi}^{text}$. $\mathcal{L}_{semi}^{merged}$ is calculated between the unlabeled prediction $y_u^{s}$ and the merged pseudo-label, where the merged pseudo-label is the combination of both the teacher network-generated pseudo-label and the text-guided mask. $\mathcal{L}_{semi}^{text}$ is calculated between the unlabeled prediction $y_u^{s}$ and the text-guided mask.

\begin{equation}
\mathcal{L}_{sup}=-\frac{1}{N_l}\frac{1}{HW}\sum_{i=1}^{N_l}\sum_{j=1}^{HW}\ell_{ce}(y_{l_{i,j}},y_{gt_{i,j}})
\end{equation}

\begin{equation}
\mathcal{L}_{semi}^{merged}=-\frac{1}{N_u}\frac{1}{HW}\sum_{i=1}^{N_u}\sum_{j=1}^{HW}\ell_{ce}(y^s_{u_{i,j}},\sigma(y^t_{u_{i,j}}+y^{text}_{u_{i,j}})
\end{equation}

\begin{equation}
\mathcal{L}_{semi}^{text}=-\frac{1}{N_u}\frac{1}{HW}\sum_{i=1}^{N_u}\sum_{j=1}^{HW}\ell_{ce}(y^s_{u_{i,j}},y^{text}_{u_{i,j}})
\end{equation}

\begin{equation}
\mathcal{L}_{semi}=(\mathcal{L}_{semi}^{merged}+\mathcal{L}_{semi}^{text})/2
\end{equation}
where $\ell(\cdot)$ is the cross-entropy loss. $(i,j)$ represents the $j$-th pixel in $i$-th mask. $N_l$ and $N_u$ are the batch size of labeled images and unlabeled images. $W$ and $H$ represent the width and height of an image.

\textbf{Summarized Advantages:} To the best of our knowledge, this is the first attempt to integrate text-guided mask into SSMIS. It compensates quality deficiencies of pseudo-labels, achieving reliable consistency regularization in SSMIS.

\subsection{Theoretical Analysis}
Our theoretical proofs have proven that SSS can address the alignment uncertainty between image and text.

\textit{\textbf{Proof.}}
\begin{equation}
\begin{aligned}
    s\widehat{i}m(I,T)&=1-(1-sim(I,T)) \cdot D_{SSS}(I,T) \\
    &= 1-(1-sim(I,T)) \cdot e^{-\lambda{\frac{D_{u}(x_1,x_2)}{D_{s}(x_1,x_2)}}} \\ 
    &\propto D_{u}(x_1,x_2).
\end{aligned}
\end{equation}

For the image-text pair with high uncertainty $D_{u}(x_1,x_2)$, i.e. high relative uncertainty $\frac{D_{u}(x_1,x_2)}{D_{s}(x_1,x_2)}$, the proposed uncertainty cosine similarity make such image-text pair more similar, addressing the alignment uncertainty.

\section{Experiments}
\label{sec4}

\subsection{Datasets}
Extensive experiments are conducted on three public medical image segmentation datasets with different tasks: chest infection area segmentation for COVID-19, bone metastases segmentation, and nuclei instances segmentation.

\textbf{QaTa-COV19} \cite{ref55} contains 9258 COVID-19 chest X-ray radiographs. \cite{ref36} provided text annotations and split 7145 samples for training and 2113 samples for testing. 

\textbf{BM-Seg} \cite{ref57} consists of 23 CT-scans from 23 patients, totaling 1517 slices from different skeletal views. 270 slices from a single view are selected, allocating 200 for training and 70 for testing. The texts are annotated by professionals.

\textbf{MoNuSeg} \cite{ref58} includes 44 images with annotations, and the image size is 1000$\times$1000. The training dataset contains 30 images and the testing dataset contains 14 images.

% Covid19---------------------------------------------------------------------------
\begin{table*}[!t]
  \centering
  \resizebox{0.9\linewidth}{!}{
   \renewcommand{\arraystretch}{1.3}
    \begin{tabular}{c|cc|cc|cc|cc|cc}
    \hline
    \multirow{2}{*}{Method} & \multicolumn{2}{c|}{Data used} & \multicolumn{2}{c|}{QaTa-COV19} & \multicolumn{2}{c|}{BM-Seg} & \multicolumn{2}{c|}{MoNuSeg} & \multicolumn{2}{c}{Complexity}  \\
\cline{2-11}          & \multicolumn{1}{c|}{Labeled} & Text & Dice (\%)  & mIoU (\%)  & Dice (\%)  & mIoU (\%) & Dice (\%)  & mIoU (\%) & Param (M)  & Flops (G) \\
    \hline
    U-Net & 100\% & $\times$    & 79.02 & 69.46 & 74.15 & 60.32 & 78.66 & 68.46 & 14.75 & 25.19 \\
    CLIP & 100\% & \checkmark     & 79.81 & 70.66 & 74.49 & 59.56 & 79.79 & 68.27 & 87.00 & 57.60 \\
    ViLT & 100\% & \checkmark     & 79.63 & 70.12 & 73.23 & 58.22 & 77.92 & 67.81 & 87.40 & 
    28.00 \\
    \hline
    MT & \multirow{9}{*}{25\%} & $\times$ & 77.88 & 67.53 & 66.31 & 53.35 & 72.80 &  56.70 & 14.75 & 25.19 \\
    CCT &       &   $\times$    & 78.02 & 67.03 & 70.66 & 55.27 & 74.25 & 56.55 & 4.65 & 8.45 \\
    BCP     &       &    $\times$   & 74.79 & 65.26 & 71.26 & 55.28  & 72.06 & 54.69 & 1.81 & 2.28 \\
    MC-Net     &       &   $\times$    & 74.58 & 64.26 & 69.67 & 53.36 & 71.53 &  48.12 & 1.81 & 2.28 \\
    SS-Net     &       &   $\times$    & 67.93 & 58.13 & 70.21 & 54.68 & 71.80 & 47.80 & 1.81 & 2.28 \\
    UCMT     &       &   $\times$    & 76.09 & 64.13 & 71.22 & 55.82 & 75.53 & 54.82 & 1.81 & 2.30 \\
    LViT     &       &   \checkmark    & 78.12 & 66.75 & 69.45 & 54.26 & 75.69 & 56.14 & 29.72 & 27.08 \\
    CMITM     &       &   \checkmark    & 78.04 & 65.84 & 70.63 & 54.66 & 75.14 & 54.28 & 14.75 & 25.19 \\
    ASG     &       &   \checkmark    & 77.92 & 65.09 & 70.26 & 54.33 & 75.69 & 55.92 & 14.75 & 25.19 \\
    MGCA     &       &   \checkmark    & 78.17 & 67.03 & 70.19 & 53.67 & 76.14 & 56.22 & 14.75 & 25.19 \\
    Ours     &       &   \checkmark    & \textbf{79.00} & \textbf{68.21} & \textbf{71.48} & \textbf{55.97} & \textbf{76.51} & \textbf{58.08} & 14.75 & 25.19 
    \\
    \hline
    MT     & \multirow{7}{*}{50\%} & $\times$ & 79.64 & 71.88 & 72.31 & 56.99 & 75.15 & 59.64 & $-$ & $-$ \\
    CCT     &       &   $\times$    & 80.25 & 71.69 & 72.65 & 60.54 & 76.23 &  55.87 & $-$ & $-$ \\
    BCP    &       &   $\times$    & 75.57 & 65.31 & 74.21 & 61.25 & 72.17 & 50.88 & $-$ & $-$ \\
    MC-Net     &       &    $\times$   & 74.98 & 61.97 & 73.37 & 60.74 & 71.59 & 48.75 & $-$ & $-$ \\
    SS-Net     &       &   $\times$    & 68.37 & 56.73 &  73.59 & 60.89 & 73.05 & 49.88 &  $-$ & $-$ \\
    UCMT     &       &   $\times$    & 77.81 & 68.65 & 74.32 & 60.41 & 76.36 & 59.29 & $-$ & $-$ \\
    LViT     &       &   \checkmark   & 80.32 & 72.16 & 73.15 & 60.14 & 76.57 & 65.44 & $-$ & $-$ \\
    CMITM     &       &   \checkmark   & 81.33 & 72.84 & 74.96 & 60.18 & 77.63 & 66.31 & $-$ & $-$ \\
    ASG     &       &   \checkmark   & 80.47 & 71.84 & 74.22 & 59.47 & 76.79 & 65.91 & $-$ & $-$ \\
    MGCA     &       &   \checkmark   & 81.24 & 73.56 & \textbf{75.17} & \textbf{61.49} & 77.06 & 65.23 & $-$ & $-$ \\
    Ours  &       &   \checkmark    & \textbf{82.52} & \textbf{75.87} & 74.61 & 61.08 & \textbf{78.03} & \textbf{66.93} & $-$ & $-$ \\
    \hline
    \end{tabular}}
  \label{table1}%
  \caption{The comparative experiments on the \textbf{QaTa-COV19}, \textbf{BM-Seg} and \textbf{MoNuSeg} datasets demonstrate that our powerful uncertainty processing and pseudo-label enhancing ability.}
\end{table*}%

% ablation study
\begin{table*}[!t]
  \centering
  \resizebox{0.7\linewidth}{!}{
    \begin{tabular}{ccccccccc}
    \hline
    \multicolumn{3}{c}{Method} & \multicolumn{2}{c}{QaTa-COV19} & \multicolumn{2}{c}{BM-Seg} & \multicolumn{2}{c}{MoNuSeg} \\
    \hline
    SSS & DCL & $\mathcal{L}_{tg}$ & Dice (\%) & mIoU (\%) & Dice (\%)  & mIoU (\%)  & Dice (\%)   & mIoU (\%) \\
    \hline
    $\times$     & $\times$     & $\times$   & 79.64  & 71.88   & 72.31   & 56.99   & 75.15   & 59.64 \\
    \checkmark     & $\times$   & $\times$   & 80.86   & 72.33   & 74.34   & 60.22   & 77.25   & 63.81 \\
    $\times$     & \checkmark     & $\times$   & 80.13   & 72.31   & 73.11   & 60.14   & 76.90   & 62.73 \\
   \checkmark     & $\times$     & \checkmark   & 81.49   & 75.25   & 75.58   & 60.79   & 77.53   & 65.10 \\
    $\times$     & \checkmark     & \checkmark   & 81.05   & 73.80   & 74.27   & 60.65   & 77.10   & 64.01 \\
    \checkmark     & \checkmark     & $\times$   & 81.10   & 75.08   & 75.24   &  59.11  & 77.74   & 64.62 \\
    \hline
    \checkmark     & \checkmark     & \checkmark   & \textbf{82.52}   & \textbf{75.87}   & \textbf{76.41}   & \textbf{61.08}   & \textbf{78.03}   & \textbf{66.93} \\
    \hline
    \end{tabular}}
  \label{table2}%
  \caption{Ablation studies demonstrate that significant improvements of the proposed innovations. The results are based on 50\% labeled data on the QaTa-COV19, BM-Seg and MoNuSeg datasets. Baseline: Teacher-Student Network; SSS: Semantic Similarity Supervision; DCL: Dual Contrastive Learning; $\mathcal{L}_{tg}$: Text-Guided Loss.}
\end{table*}%

\subsection{Implementation Details}
Our method is implemented using Pytorch. The operating system is Ubuntu 20.04.4 LTS with 24GB V100 GPU. The learning rate is set to 3e-4 for both the QaTa-COV19 and BM-Seg datasets, and 1e-3 for the MoNuSeg dataset. Early stopping is implemented if the model's performance does not improve after 20 epochs based on its current performance. The batch size is 32 for the QaTa-COV19 dataset, 16 for the BM-Seg dataset, and 4 for the MoNuSeg dataset.

\begin{figure*}[t]
\centering
\includegraphics[width=0.8\linewidth]{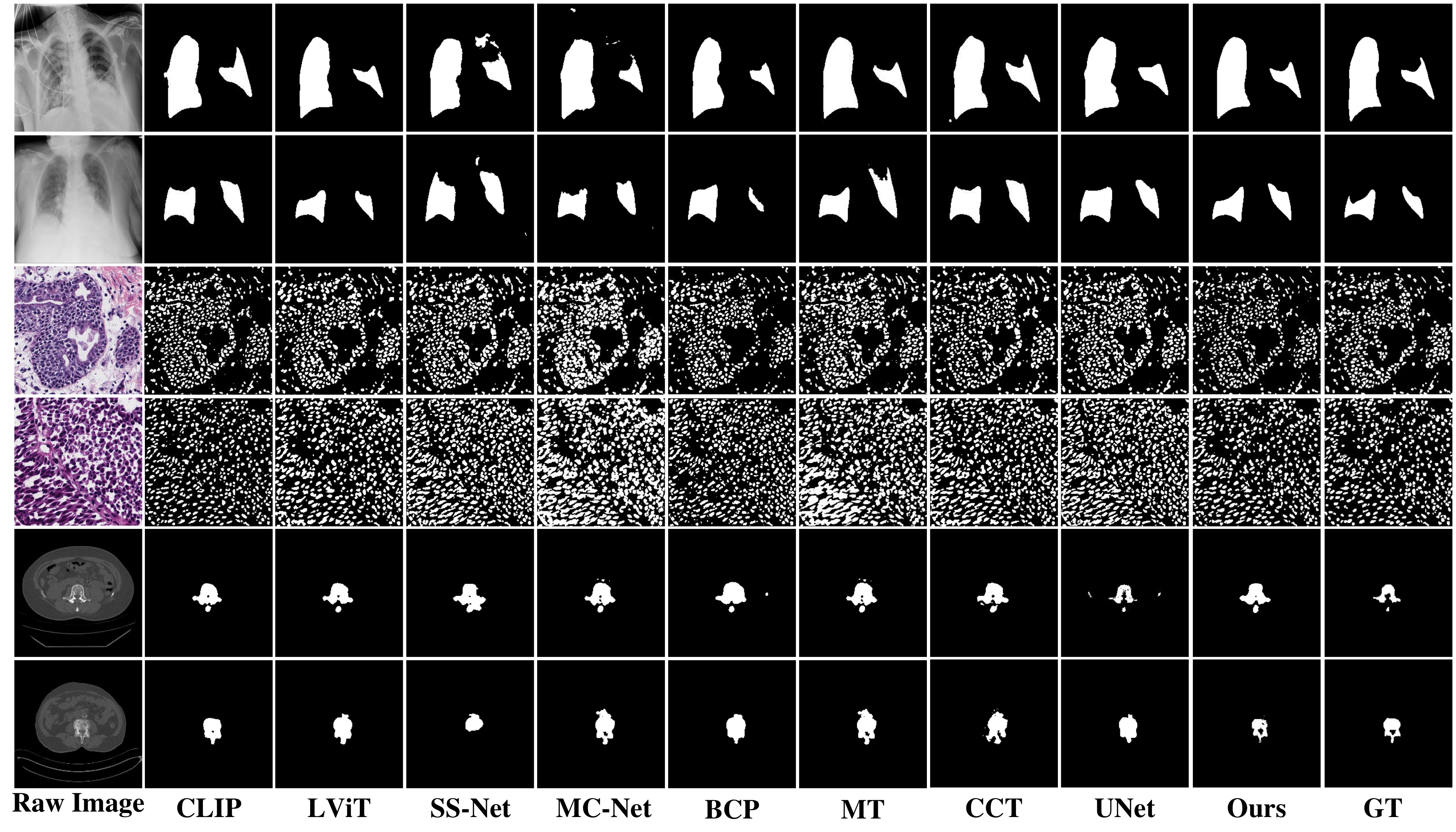}
\caption{The visual superiority of the proposed method (DuSSS) on the QaTa-COV19, BM-Seg and MoNuSeg datasets. The proposed method shows high-quality segmentation, compared with semi-supervised and VLM-based methods.}
\label{fig4}
\end{figure*}

\subsection{Evaluation Metrics}
The Dice $Dice=\sum_{i=1}^N\sum_{j=1}^C\frac{1}{NC}\cdot\frac{2|p_{ij}\cap y_{ij}|}{(|p_{ij}|+|y_{ij}|)}$ and mIoU $mIoU=\sum_{i=1}^N\sum_{j=1}^C\frac{1}{NC}\cdot\frac{|p_{ij}\cap y_{ij}|}{|p_{ij}\cup y_{ij}|}$ are used to evaluate our method and other compared methods, where $C$ is the number of categories, and $N$ is the number of pixels.

\subsection{Comparison Study Shows Our Superiority}
The DuSSS is compared on the QaTa-COV19, BM-Seg and MoNuSeg datasets with 13 state-of-the art methods, including U-Net \cite{ref59}, CLIP \cite{ref25}, ViLT \cite{ref26}, MT \cite{ref62}, CCT \cite{refa26}, BCP \cite{ref64}, MC-Net \cite{ref65}, SS-Net \cite{ref66}, UCMT \cite{ref67}, and LViT \cite{ref36}, CMITM \cite{refa28}, ASG \cite{refa29}, MGCA \cite{refa15}.

\textbf{Qualitative Analysis.}
Fig. \ref{fig4} presents the excellent segmentation performance of our method in comparison with state-of-the art methods. It can be observed that our method not only accurately localizes target regions but also generates coherent boundaries, even in small object circumstances. As shown in Fig. \ref{fig4}, SS-Net, MC-Net, BCP, MT and CCT all have more severe mis-segmentation than the proposed method. This indicates that the introduction of the text-guided mask in the semi-supervised learning can better guide the training of the model, and consequently generate more accurate segmentation. In addition, compared with different VLM-guided segmentation methods, the DuSSS is more delicate in the segmentation boundary. This is attributed to the understanding of cross-modal semantic uncertainty during the pre-training process.

\textbf{Quantitative Analysis.} 4
Our DuSSS achieves highly competitive performance on three datasets against existing state-of-the art methods. Specifically, Table 1 indicates that our DuSSS outperforms all the other methods in both Dice (82.52\%) and mIoU (75.87\%) on 50\% labeled images, and both Dice (79.00\%) and mIoU (68.21\%) on 25\% labeled images from the QaTa-COV19 dataset. It is noteworthy that the proposed method surpasses fully supervised methods. Additionally, the proposed method achieves better performance compared to the text-equipped methods, while requiring fewer parameters and having lower computational costs. It indicates that our DuSSS driven SSMIS has the ability of precisely locating target regions with text guidance. Table 1 also demonstrates the superiority our DuSSS in the BM-Seg and MoNuSeg datasets. Overall, our DuSSS achieves the best results in different segmentation tasks, demonstrating its superior generalization and robustness.

\begin{figure}[h]
\centering
\includegraphics[width=\linewidth]{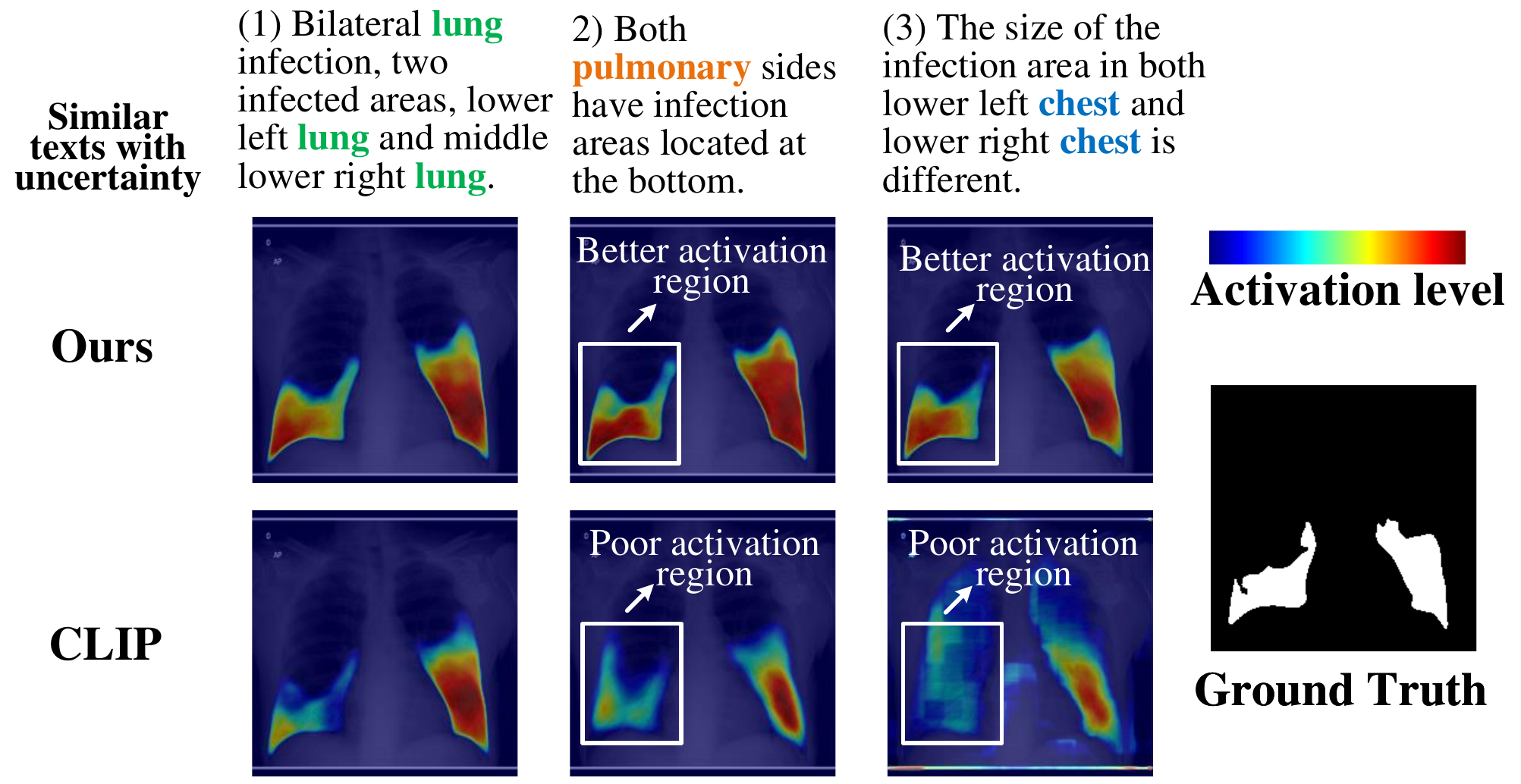}
\caption{Our DuSSS effectively addresses the uncertainty problems resulting from the diverse correspondences between image and text. The DuSSS shows great target region activation effects for multiple similarly yet lexically ambiguous texts, highlighting its powerful robustness.}
\label{fig5}
\end{figure}

\subsection{Ablation Study}
We perform ablation experiments to verify the contribution of novel designs, and the number of segmentation model parameters does not increase after adding novel designs.

\textbf{Effectiveness of SSS.} Table 2 shows that removing the SSS limits the performance, resulting in a decrease of 1.22\% in Dice and 0.45\% in mIoU on the QaTa-COV19 dataset. Additionally, adding SSS to DCL could further improve 0.97\% in Dice and 2.77\% mIoU. This indicates that the SSS contributes to supervising semantic similarity in ambiguous data via being aware of the uncertainty.

\textbf{Effectiveness of DCL.}
Table 2 demonstrates that the DCL improves the Dice and mIoU by 0.49\% and 0.43\% on the QaTa-COV19 dataset respectively, compared with the cross-modal contrastive learning alone. This owes to alleviating uncertainty during image-text alignment by enhancing semantic consistent representations within each modality.

\textbf{Effectiveness of Text-Guided Loss.}
Table 2 indicates that introducing the text-guided loss improves the performance of SSS, DCL, and their combination. It demonstrates the superiority of text-guided loss in restraining the interference of negative text regions.

\subsection{Uncertainty Analysis}
Fig. \ref{fig5} demonstrates the strong robustness of our DuSSS against uncertainty problems. For multiple similar texts with lexical uncertainty, our DuSSS generates robust target region activation effects. However, other VLM-based method shows unstable activation effects due to being unaware of uncertainty. Therefore, the proposed DuSSS can comprehend various semantic correspondences between image and text, aiding the model to better address semantic uncertainty, thus enhancing its performance in the image-text alignment.

\section{Conclusion}
\label{sec6}
In this paper, we propose a novel VLM named DuSSS for SSMIS. In VLM pre-training, the DCL reinforces semantic correlations by learning intrinsic representations within each modality. Moreover, the DuSSS integrates the SSS into each contrastive learning process to supervise semantic similarity based on uncertainty levels, addressing semantic uncertainty across modalities. Finally, using the pretrained VLM, a text-guide SSMIS framework is proposed to enhance the quality of pseudo labels, improving the model’s consistency learning capability. Experimental results demonstrate that our DuSSS outperforms SOTA methods.

\section{Acknowledgments}
\label{sec7}
This work was partly supported by the National Natural Science Foundation of China (Grant No.62173212), Taishan Scholars Program of Shandong Province(Grant No.tsqn202306017), Shandong Province“Double-Hundred Talent Plan”on 100 Foreign Experts and 100 Foreign Expert Teams (Grant No.WSR2023049).

\bibliography{aaai25}

\end{document}